\documentclass[final]{AISB2008}

\usepackage{times}
\usepackage{graphicx}
\usepackage{latexsym}
\usepackage{hyperref}
\usepackage{listings}
\usepackage{booktabs}
\usepackage{subcaption}

\begin{document}

\title{A Practical Guide to Studying Emergent Communication through Grounded Language Games}
\author{Jens Nevens$^1$ \and Paul Van Eecke$^1$ \and Katrien Beuls\institute{Artificial Intelligence Lab, Vrije Universiteit Brussel, Belgium, correspondence: \href{mailto:ehai@ai.vub.ac.be}{ehai@ai.vub.ac.be}} \\ \\ \normalfont{\textit{This paper was officially published at the}} \\ \normalfont{\textit{`Language Learning for Artificial Agents (L2A2) Symposium'}} \\ \normalfont{\textit{of the 2019 Artificial Intelligence and Simulation of Behaviour (AISB) Convention.}}}

\maketitle
\bibliographystyle{AISB2008}

\begin{abstract}
The question of how an effective and efficient communication system can emerge in a population of agents that need to solve a particular task attracts more and more attention from researchers in many fields, including artificial intelligence, linguistics and statistical physics. A common methodology for studying this question consists of carrying out multi-agent experiments in which a population of agents takes part in a series of scripted and task-oriented communicative interactions, called `language games'. 
While each individual language game is typically played by two agents in the population, a large series of games allows the population to converge on a shared communication system. 
Setting up an experiment in which a rich system for communicating about the real world emerges is a major enterprise, as it requires a variety of software components for running multi-agent experiments, for interacting with sensors and actuators, for conceptualising and interpreting semantic structures, and for mapping between these semantic structures and linguistic utterances. The aim of this paper is twofold. On the one hand, it introduces a high-level robot interface that extends the Babel software system, presenting for the first time a toolkit that provides flexible modules for dealing with each subtask involved in running advanced grounded language game experiments. On the other hand, it provides a practical guide to using the toolkit for implementing such experiments, taking a grounded colour naming game experiment as a didactic example.
\end{abstract}

\section{INTRODUCTION}
\label{sec:introduction}

%meer interesse (NIPS community), long history, 
How can a population of agents self-organise an effective and efficient communication system that allows them to communicate about their native environment? This fundamental research question concerning the mechanisms underlying human-like communication systems has for a long time sparked the interest of researchers from many fields, including artificial intelligence (e.g. \cite{steels1995self,mordatch2017emergence}), linguistics (e.g. \cite{van2013linguistic,lestrade_2016}), and statistical physics (e.g. \cite{baronchelli2006sharp,puglisi2008cultural}).  Well-attended workshops at important conferences, such as the NeurIPS emergent communication workshop, indicate that the community interested in models of emergent communication is growing ever more rapidly.

%why robots, language evolution/emergent communication  (NIPS papers?, Experiments book.... Spranger, Pauw, Loetzsch, Steels ....
A common methodology for studying emergent communication consists of carrying out multi-agent experiments in which a population of agents takes part in a series of scripted and task-oriented communicative interactions, called `language games' \cite{steels1995self}. Each game is typically played locally by two agents in the population without any form of central control and without the agents having any mind-reading capacities. Through self-organisation, the population converges on a shared communication system after playing a large number of games \cite{de2006reach}. The most widely studied language game is the Naming Game \cite{steels1995self, steels1998spatially, baronchelli2006sharp}, in which the task involves referring to individual objects and thereby establishing a shared lexicon of proper nouns. More advanced scenarios include games in which the agents refer to the properties of objects or events \cite{van2008emergence}, use multi-word expressions \cite{van1999multiple}, or develop grammatical structures \cite{beuls2013agent}.

Mathematical investigations and computer simulations help making the assumptions of a specific theory explicit, allowing researchers to study the emergence of a particular communication system in a simulated world, comparing different scenarios and parameter settings. Yet, the step from such a simulated world to the real world with noisy sensori-motor values is crucial to make and requires the use of physical robots, as has been advocated in the work of many researchers studying the emergence and evolution of speech and language \cite{loetzsch10why, language-grounding-in-robots, oudeyer2006discovering}. The increased realism leads to the need for more robust and fine-grained models, as has for instance been shown when moving from the Naming Game in a simulated world to the Grounded Naming Game in the real world \cite{steels2012grounded, steels2016boy}.

\begin{figure}[t]
\includegraphics[width=\columnwidth]{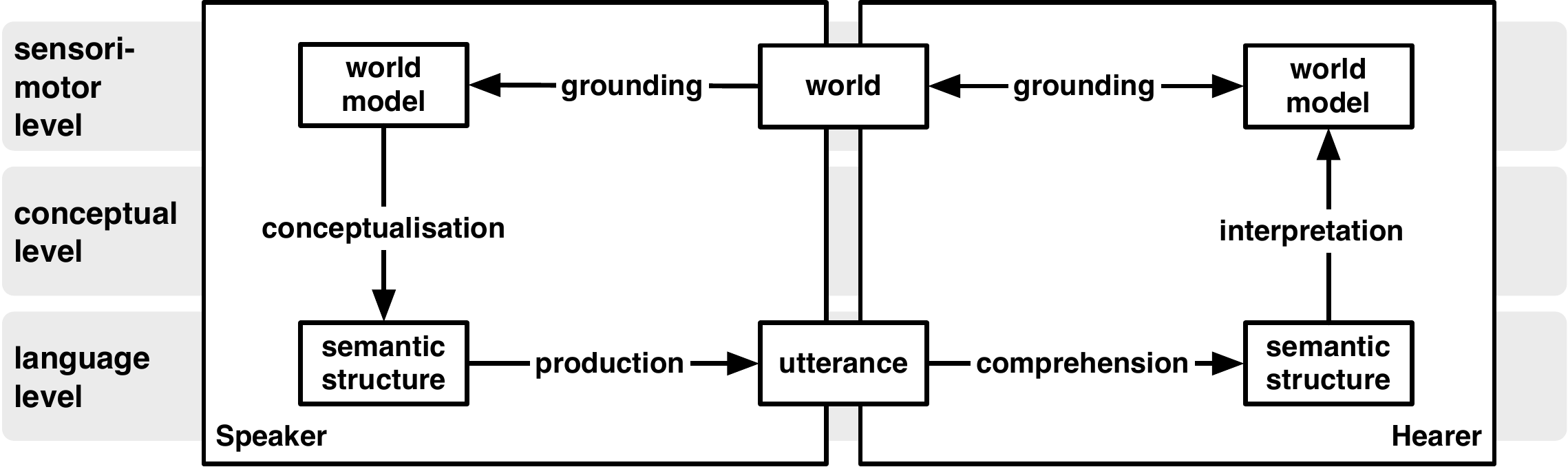}
\caption{Grounded robot interactions follow a semiotic cycle that involves three main levels: the sensori-motor level, the conceptual level and the language level. The speaker agent (on the left hand-side) and the hearer agent (on the right hand-side) only share the world in which they are situated and the utterance that the speaker produces.}
\label{fig:semiotic-cycle}
\end{figure}

%requirements: semiotic cycle - eigenlijk moeten we language games nog introduceren??
Setting up such grounded language game experiments requires taking into consideration a set of processes that has been referred to as \textit{the semiotic cycle} \cite{steels2012grounding}, and implementing each of the processes involved. During each game, the speaker and hearer move through this cycle as depicted in Figure \ref{fig:semiotic-cycle}. First, both agents perceive the world through their own sensors and construct an internal world model (\textit{grounding}). Then, the speaker determines which information needs to be conveyed to the hearer in order for the task to succeed and conceptualises it into a semantic structure (\textit{conceptualisation}). This meaning representation is then expressed through a linguistic expression that is passed to the hearer (\textit{production}). The hearer then parses the utterance into a meaning representation (\textit{comprehension}). He interprets the resulting semantic structure in relation with his world model and performs the relevant action (\textit{interpretation}). Finally, the speaker provides feedback on the outcome of the game, allowing both agents to update their individual knowledge. The described processes take place on three main levels: grounding on the sensori-motor level, conceptualisation and interpretation on the conceptual level, and production and comprehension on the language level.

%weinig tools beschikbaar, verwijzen naar Netlogo voor multi-agent experiments, MOLE (Lestrade), Schueller ()
%Babel (manual, angus babel1, language grounding in robots)
There are a number of tools available that can be used for implementing language game experiments. A general-purpose, widely used platform is \textit{NetLogo} \cite{netlogo}, which was developed as an educational tool to teach students about agent-based modelling. It is mainly targeting the complex systems science community and contains a large number of sample models on many topics. NetLogo provides an excellent architecture for setting up and monitoring multi-agent simulations but does not contain any built-in functionalities for implementing the processes involved in the semiotic cycle. 

\textit{NaminggamesAL} is a recent tool for implementing a variety of basic naming games in simulated worlds \cite{schueller2018}. It includes a multi-agent simulation framework and a number of built-in learning strategies, but lacks modules  for implementing more advanced versions of the three levels of the semiotic cycle.

A software tool that stems from the linguistic community is \textit{MoLE} (\textit{Modelling Language Evolution}) \cite{mole}. MoLE focuses on the language level and was especially designed for conducting experiments on the emergence of case \cite{lestrade_2016}. It includes the necessary building blocks for setting up multi-agent language games in which lexical items can be recruited as grammatical markers. It does not include an advanced semantic processing engine, an elaborate language processing engine, and interfaces to physical robots or rich world models. 

Finally, \textit{Babel} is a software package that was originally implemented as a testbed for research on the origins of language \cite{McIntyre:1998}. In its first version, it provided users with a basis for running computer simulations and allowed the rapid construction of experiments and a flexible visualisation of the results. Later versions of Babel (see \cite{steels2010babel, loetzsch2008babel2}) introduced more elaborate tools for setting up language games with advanced modules for dealing with the conceptual level (IRL -- \cite{steels00e, spranger2012irl}) and the language level (FCG -- \cite{steels2011design, steels2017basics}). Although Babel has often been used in grounded experiments, involving amongst others the AIBO dog-like robot \cite{steels2000aibo}, the QRIO humanoid \cite{spranger12perceptual} and the PERACT vision system \cite{vantrjip2016}, it has never included a standardised interface to connect to robotic platforms.

The contribution of this paper is twofold. On the one hand, it introduces a high-level robot interface that extends the Babel software system, presenting for the first time a toolkit that provides flexible modules for dealing with each subtask involved in running advanced grounded language game experiments. On the other hand, it provides a practical guide to using the toolkit for implementing such experiments, taking a grounded colour naming game experiment as a didactic example. 

The remainder of this paper is structured as follows. Section \ref{sec:colour} introduces the challenges involved in establishing a shared colour lexicon and discusses the grounded colour naming game as a solution. Section \ref{sec:implementing-a-game} serves as a practical explanation of how this solution can be implemented on a high level using the Babel system. Finally, Section \ref{sec:robot-interface} provides more technical detail on the architecture and main features of the newly developed robot interface.

\section{EMERGENT COMMUNICATION FOR THE COLOUR DOMAIN}
\label{sec:colour}
%JORIS, tony, ...
%probleem introduceren, moeilijkheden, foto met knuffels
%rgb niet doorsturen
The goal of the colour naming game experiment is to show how a shared communication system for referring to objects by their colour can emerge in a population of autonomous agents. The agents start without any concepts or words, perceiving only the average colour values of the objects in the scene. In a real-world setting, transmitting raw sensor values does not lead to successful communication, because the sensors of each agent will always record slightly different values due to differences in the agents' perspectives on the scene, changes in lighting conditions and in some cases differences in robot morphology\footnote{Traditional sensor calibration is undesirable here, as it requires a notion of central control which conflicts with the autonomous nature of the agents.}. Therefore, concepts and words form the necessary layers to abstract away from sensor data, in order to achieve more robust communication.

%hoe wordt het opgelost: literatuur
A large body of previous work has shown how colour categories and words can emerge through self-organisation in a population of autonomous agents, including robots \cite{steels2005coordinating,puglisi2008cultural,bleys2009grounded, bleys2016language,cornudella2016role}. In essence, the solution resides in the agents dividing their continuous colour space into convex regions that correspond to colour categories that are functional in the world, and in establishing a shared lexicon to refer to each region. An operationalisation of this solution has been proposed in the form of \textit{the grounded colour naming game experiment} \cite{steels2005coordinating}. 

\begin{figure}
\centering
\includegraphics[width=\columnwidth]{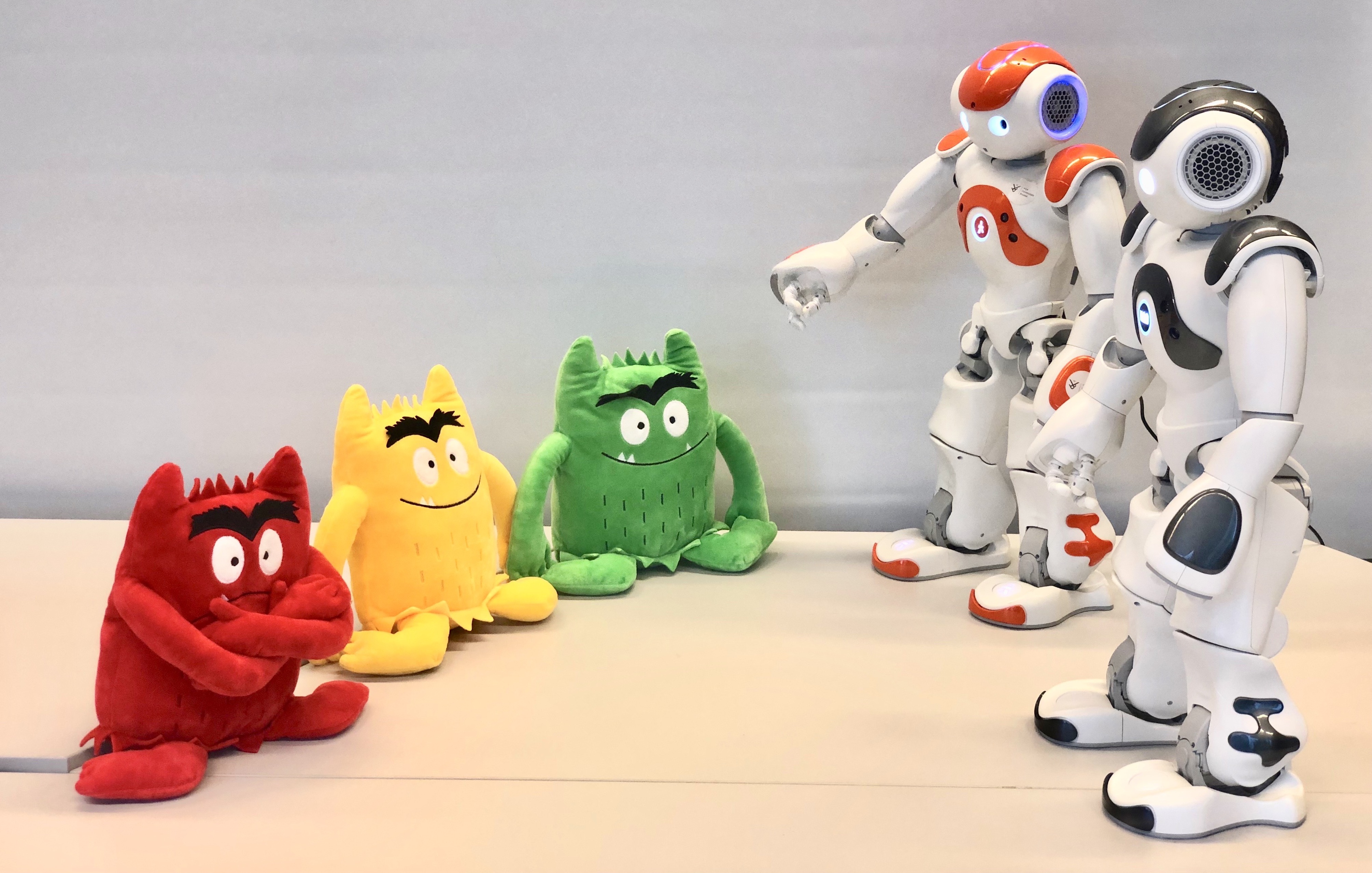}
\caption{Two Nao robots play a grounded colour naming game with a scene consisting of coloured monsters. For each game, two agents from the population are physically instantiated in the two robot bodies.}
\label{fig:naos}
\end{figure}

%game en dynamica
Figure \ref{fig:naos} shows an instantiation of a grounded colour naming game. In this game, the world consists of a number of toy monsters, each with a different colour. Two randomly selected agents from the population are physically embodied in the two robots, one playing the role of speaker and the other the role of hearer. The task of the speaker is to use a vocalisation to draw the attention of the hearer to one of the monsters in the world. The task of the hearer is to point to this monster, signalling that he has understood the message. Finally, the speaker signals success if the hearer pointed to the right monster, or points himself if this was not the case.

%invention, adoption, alignment
Once the basic interaction script is in place, we can start experimenting with different mechanisms for inventing, adopting and aligning colour categories and words. Suppose that the orange robot in the back of Figure \ref{fig:naos} needs to refer to the green monster in front of him. As he enters the experiment without any colour categories or words, he needs to invent both a new category and a new word to express this category. He takes the observed colour value as the first prototypical value of this new category (e.g. [(7, 246, 9) $\leftrightarrow$ \texttt{CATEGORY-1}]) and associates the category with a newly generated word form, in this case ``fusemo", assigning to the association a default initial score (e.g. [\texttt{CATEGORY-1} $\leftrightarrow$ \textit{fusemo}]). He then utters the word to the hearer. The hearer does not know this newly invented word and is therefore unable to determine which monster the speaker is referring to. The speaker provides feedback by pointing to the green monster. At that point, the hearer associates the colour value that he observed for this object to a new category (e.g. [(5, 243, 2) $\leftrightarrow$ \texttt{CATEGORY-2}]) and associates this category to the word ``fusemo" with a default initial score (e.g. [\texttt{CATEGORY-2} $\leftrightarrow$ \textit{fusemo}]). Crucially, each association between a sensor value and a category, as well as between a category and a word form is internal to the individual agent and cannot be shared as such.

%als je al een categorie hebt - DISCRIMINATIE
While agents are able to invent new categories and words throughout the experiment, they will prefer to reuse existing ones. When a novel observation comes in, the agents will determine whether the category that is the closest in sensory distance to the observation discriminates the topic monster from the other monsters in the world. In other terms, they will calculate the distance between the observed sensory value and each of their categories, and select the closest one. Then, they verify that no other object is closer to this category, which means that the category uniquely discriminates the monster in the world. If no such category can be found,  a new category is invented following the procedure explained in the previous paragraph. When it comes to the words, the speaker will always choose the word form most strongly associated with the selected category and the hearer will choose the category most strongly associated with the selected word form.

%alignment - scores en shifting
When agents invent, adopt and reuse colour categories and words as described above, the categories of the agents never align and their vocabularies become enormous in size. To overcome this issue, speaker and hearer go through an alignment phase at the end of each interaction. If the interaction succeeds, the agents reinforce the association between the category and the word form that was used and punish competing associations. Moreover, they also slightly shift the prototypical value of the used category towards the observed sensor value. If the game fails, the agents punish the association that was used.

%conclusie en hoe kan je dit nu implementeren?
Using these mechanisms for invention, adoption and alignment, a population of agents will eventually converge on a stable colour category system, and on a shared inventory of words for referring to these categories. Importantly, the emerged system is tailored towards the distinctions that are functional in the world, both in terms of number of categories and in the way in which the colour space is subdivided. The concrete mechanisms described in this section are the ones most commonly used in the literature. A complete overview of invention, adoption and alignment strategies  that have been explored for the colour naming game falls outside the scope of this paper, but can be found in earlier work by Joris Bleys \cite{bleys2016language}.

\section{IMPLEMENTING A GROUNDED COLOUR NAMING GAME EXPERIMENT}
\label{sec:implementing-a-game}

We will now demonstrate how the Babel toolkit can be used to implement a grounded colour naming game experiment like the one that was introduced in the previous section. Babel's experiment framework and submodules for dealing with the sensori-motor level, conceptual level and language level provide abstractions that allow specifying the game on a high level and in an intuitive way, while most technical detail is taken care of by the system itself \footnote{In this example experiment, all processes in the semiotic cycle are implemented using standard Babel modules. It is however perfectly possible to only use Babel modules for implementing some of these processes, and different software for the others.}. Actual source code that corresponds to the explanation in this section has become an integral part of the Babel toolkit\footnote{The complete source code for running the Grounded Colour Naming Game Experiment is part of the Babel toolkit and can be found in the subfolder \texttt{experiments/grounded-colour-naming-game-experiment}. A simulator mode has also been provided, to run the experiments if you do not have a Nao robot at your disposal.}, which can be obtained via \url{https://emergent-languages.org}. Additionally, an online web demonstration of the grounded colour naming game experiment is available at \url{https://ehai.ai.vub.ac.be/demos/babel-grounded-colour-naming-game-experiment}.

\subsection{Multi-agent architecture}
\label{sec:multi}

Implementing a language game experiment involves keeping track of the agents in the population, selecting the agents to participate in each game, assigning them the role of speaker or hearer, and, most importantly, specifying the language game script according to which the agents will interact. Within Babel, the multi-agent simulation part of the experiment is handled by the `experiment-framework' submodule. The experiment framework is entirely customisable when it comes to how the population is structured, which and how many agents are selected for each interaction, how their role is determined and what an interaction looks like. For this grounded colour naming game experiment, we will make use of the experiment framework's default settings: a fully connected population structure, one speaker and one hearer per game, both randomly selected, and communicative success as a measure for evaluation. The interaction script itself is specified as shown in Listing \ref{interaction-script} and consists of the following steps (with the Babel function names between parentheses):

\begin{enumerate}
\item Two agents are downloaded into the robot bodies (\textsc{embody})
\item Both agents scan the world and construct their world model (\textsc{agent-observe-world})
\item The speaker chooses an object to refer to (\textsc{choose-topic})
\item The speaker conceptualises the topic in relation to his world model (\textsc{conceptualise})
\item The speaker chooses a word for the topic (\textsc{produce-utterance})
\item The speaker utters the word while the hearer is listening (\textsc{pass-utterance})
\item The hearer parses the observed word into a semantic structure (\textsc{comprehend-utterance})
\item The hearer interprets the semantic structure in relation to his world model (\textsc{interpret})
\item The hearer points to the hypothesized topic (\textsc{point-and-observe})
\item The speaker provides feedback by nodding (\textsc{agent-nod}) in case of success, or pointing (\textsc{point-and-observe}) in case of failure
\item Both agents align (\textsc{align-agent})
\end{enumerate}

\textsc{embody}, \textsc{agent-observe-world}, \textsc{pass-utterance}, \textsc{point-and-observe} and \textsc{agent-nod} all happen at the sensori-motor level; \textsc{choose-topic}, \textsc{conceptualise} and \textsc{interpret} at the conceptual level; and \textsc{produce-utterance} and \textsc{comprehend-utterance} at the language level. Finally, \textsc{align-agent} takes place on both the conceptual and the language level.

\begin{figure}
\begin{lstlisting}[caption=Interaction Script, label=interaction-script, basicstyle=\scriptsize\tt]
(defmethod interact ((experiment gcng-experiment) interaction &key)
  (let ((speaker (speaker interaction))
        (hearer (hearer interaction)))
    ;; 1
    (embody speaker (first (robots experiment)))
    (embody hearer (second (robots experiment)))
    ;; 2
    (agent-observe-world speaker)
    (agent-observe-world hearer)
    ;; 3
    (choose-topic speaker (world speaker))
    ;; 4
    (conceptualise speaker (topic speaker) (world speaker))
    ;; 5
    (produce-utterance speaker (meaning-representation speaker))
    ;; 6
    (pass-utterance speaker hearer (utterance speaker))
    ;; 7
    (comprehend-utterance hearer (observed-utterance hearer))
    ;; 8
    (interpret hearer (meaning-representation hearer))
    ;; 9
    (point-and-observe hearer (hypothesized-topic hearer))
    ;; 10
    (if (communicated-successfully interaction)
      (agent-nod speaker)
      (point-and-observe speaker (topic speaker)))
    ;; 11
    (align-agent speaker)
    (align-agent hearer)))
\end{lstlisting}
\end{figure}

\subsection{Sensori-motor level}%actuation?
\label{sec:perception-and-action}

The agents' action and perception capabilities are handled at the sensori-motor level by Babel's `robot-interface' submodule, which is presented for the first time in this paper. The robot interface defines a standard set of functions that are particularly useful for conducting language games, for example scanning the robot's environment, speaking, listening and pointing. It abstracts away these high-level instructions from their specific implementation, which heavily depends on the hardware that is used, and is different for each type of robot. More technical detail can be found in Section \ref{sec:robot-interface} below, with an overview of the available functionality in Table \ref{tab:robot-interface}. 

In the example experiment presented in this paper, the robot interface makes use of the sensors and actuators of the Nao robotic platform. Concretely, the \textsc{embody} step embodies the speaker and hearer agents into the available robots. The \textsc{agent-observe-world} step uses the camera of the robot to make a picture of the scene, and uses the OpenCV library \cite{opencv_library} to construct a world model by segmenting the scene and extracting certain features for the objects, including their average colour value.  \textsc{pass-utterance} lets one robot speak via text-to-speech while the other listens and performs speech recognition. \textsc{point-and-observe} is used by the hearer to indicate the hypothesized object. Finally, either  \textsc{agent-nod} or \textsc{point-and-observe} is used by the speaker at the end of the game to signal success or provide feedback.

\subsection{Conceptual level}
\label{sec:conceptualisation-and-interpretation}

Bridging the gap between the world model and the meaning that needs to be expressed by the speaker or interpreted by the hearer is handled at the conceptual level by Babel's `IRL' (Incremental Recruitment Language) submodule \cite{steels00e,spranger2012irl}. IRL implements a form of procedural semantics, which means that semantic representations consist of primitive operations that directly correspond to actual function calls, and which can be combined into semantic networks for expressing more complex meanings. In conceptualisation, the IRL engine uses a search process to compose such a semantic network that singles out a given topic in the current scene. In interpretation, the IRL engine executes the semantic network by calling the functions underlying the primitive operations and propagating the resulting values.

\begin{figure}
\centering
\includegraphics[width=.75\columnwidth]{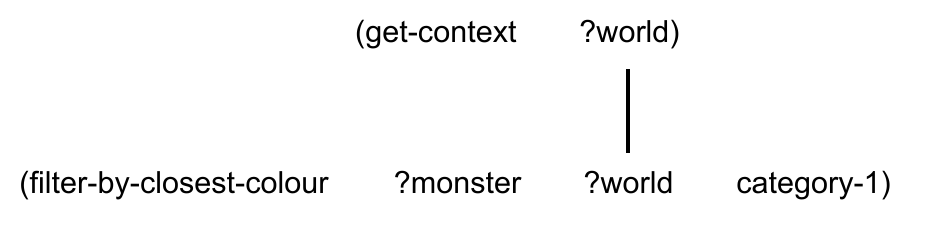}
\caption{The speaker's semantic network that singles out an object discriminating it by \texttt{CATEGORY-1}.}
\label{fig:irl-example}
\end{figure}

Suppose that in this example experiment the speaker needs to refer to the green monster. \textsc{conceptualise}  will then trigger the IRL engine to compose the smallest possible semantic network that uniquely discriminates the object by its colour, relying on the agent's ontology. As the present experiment is only concerned with basic colour categories, the semantic network will always consist of a single \textsc{filter-by-closest-colour} operation, in this case using \texttt{CATEGORY-1}, as shown in Figure \ref{fig:irl-example}. On the hearer's side, \textsc{interpret} calls the IRL engine to execute the semantic network that results from the comprehension process, also consisting here of a single \textsc{filter-by-closest-colour} operation,  in order to retrieve the topic object. During the alignment phase at the end of a successful game, the prototypical value of the used categories in the speaker's and hearer's ontologies are updated by slightly shifting them towards the values that were observed in this game.

For didactic purposes, only the very basic functionality of IRL is used in this experiment. For experiments that necessitate more complex semantic structures, we refer the reader to earlier work by Spranger (on spatial language) \cite{spranger2012grammar}, Bleys (on colour) \cite{bleys2016language} and Pauw (on quantification) \cite{pauw2012emergence}.

\subsection{Language level}
\label{sec:production-and-comprehension}

The task of mapping between a semantic structure and an utterance is taken care of by Babel's `FCG' (Fluid Construction Grammar) submodule \cite{steels2011design, steels2017basics}. FCG performs this mapping based on emergent form-meaning pairings, in this context called constructions. On the form side, a construction can include any form-related features, such as word forms, morphological properties and word order constraints. On the meaning side, it can include any type of semantic information, for example (parts of) a semantic network composed at the conceptual level.

\begin{figure}
\centering
\includegraphics[width=\columnwidth]{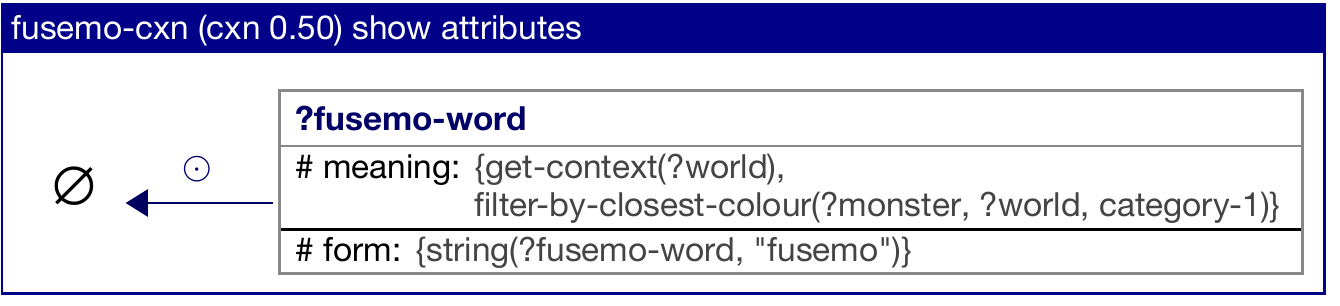}
\caption{The speaker's construction that maps between the word form ``fusemo" and its meaning.}
\label{fig:fcg-example}
\end{figure}

In the example above, the speaker invented the word ``fusemo" to refer to the colour of the green monster. He will use FCG to create a new construction that maps between this word form and the semantic network that was the outcome of the conceptualisation process. The construction is initialised with a default entrenchment score of 0.50, as illustrated by Figure \ref{fig:fcg-example}. As the hearer had never heard this word before, after feedback he will create his own construction that maps between the observed word form ``fusemo" and the semantic network that results from conceptualising in his world model the object that was pointed at. If the necessary constructions are already in place, the speaker uses \textsc{produce-utterance} to find the word form most strongly associated to his meaning network and the hearer uses \textsc{comprehend-utterance} to retrieve the meaning network most strongly associated to the word form that he observed. 

During the alignment phase after a successful game, both agents will increase the entrenchment score of the constructions that they have used. The agents will also decrease the score of competing constructions, i.e. constructions that map the same meaning to other word forms in the case of the speaker, and constructions that map the same word form to other meanings in the case of the hearer. After a failed game, both agents decrease the score of the constructions that were applied. When the score of a construction reaches zero, the construction is removed from the agent's inventory.

The constructions that are used in this didactic example are always direct mappings between a single word form and a complete meaning network. Moreover, a single construction always suffices to comprehend and produce an utterance. Examples of experiments that involve more complex linguistic structures can for example be found in previous work by Garcia Casademont (on hierarchical structures) \cite{garcia2016insight} and Beuls (on grammatical agreement) \cite{beuls2013agent}.

\subsection{Running and monitoring experiments}
%running the experiments (series/batch) en monitoring, vb. van graphs (ook plot-raw-data vermelden)

The Babel toolkit comes with a `monitors' submodule that is designed to track a multitude of experimental parameters in real time. The data recorded during a series of experiments can be displayed using dynamically updating graphs or exported to data files for later data exploration. Experimental parameters that are typically tracked include communicative success, the number of constructions in the inventories of the interacting agents, the categories in their ontologies and the size of the semantic (IRL) and syntactic (FCG) search spaces.

\begin{figure} 
\includegraphics[width=\columnwidth]{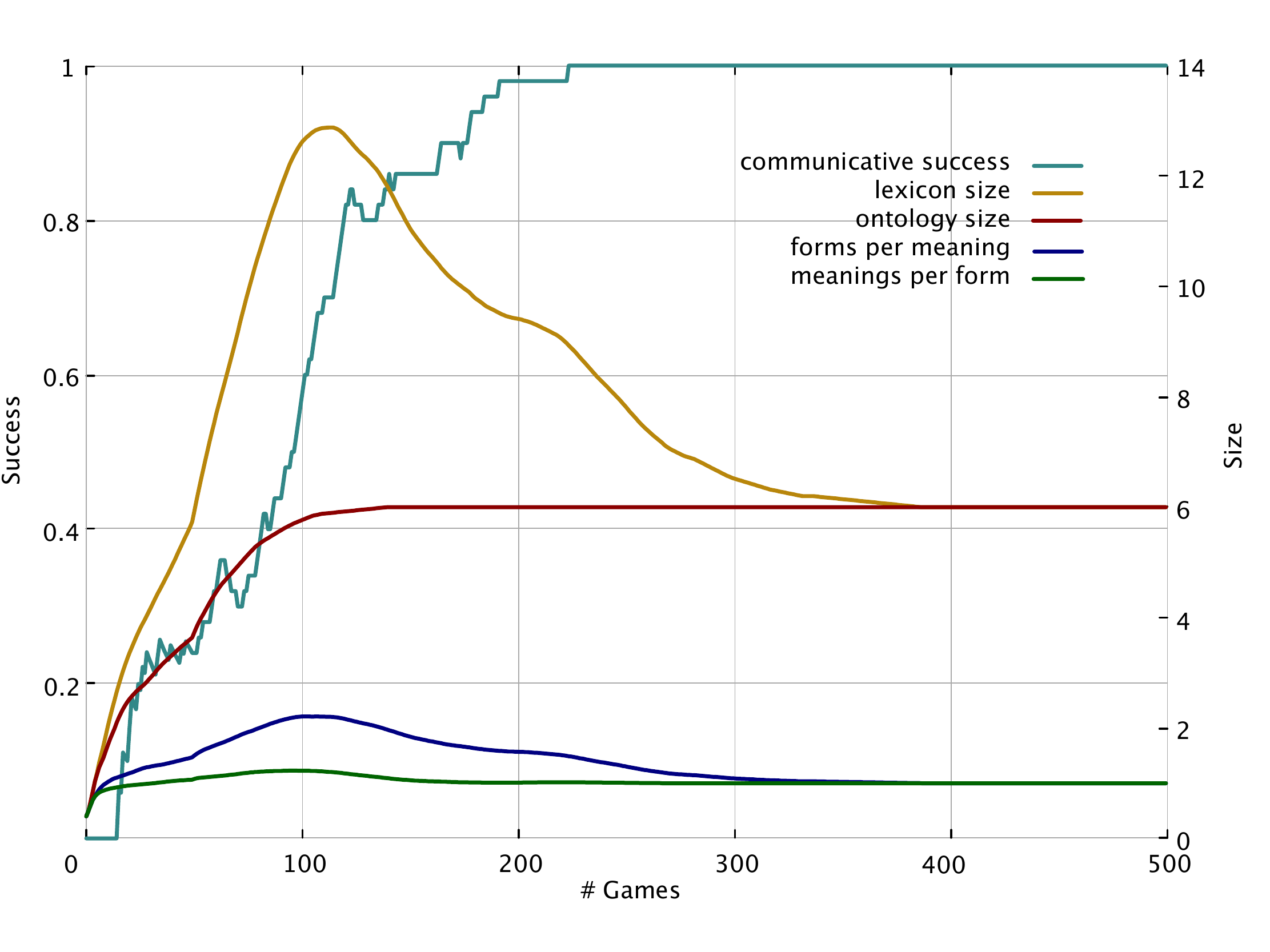}
\caption{A graph showing the results of monitoring a single experimental run of the grounded colour naming game experiment. The population consists of five agents that develop a communication system to refer to six distinctly coloured monsters.}
\label{fig:results}
\end{figure}

Figure \ref{fig:results} shows a graph that was created by the monitoring system during a single experimental run of the grounded colour naming game experiment. There were five agents in the population, communicating about six distinctly coloured monsters of which three were shown during each game. The x-axis represents the time dimension, indicating the total number of games that were played. The turquoise line indicates the average communicative success that was achieved over the last fifty games (left y-axis). At the beginning of the run, the communicative success equals zero as the agents start without any categories or words. Over the course of 250 games, it rises to 1, as the emerged communication system becomes powerful enough to solve the task. The ontology size (red line), i.e. the average number of categories per agent, starts at zero and goes to six in just over 100 games (right y-axis). This number is optimal for this experimental set-up, as there are indeed six colour distinctions that are useful in the world. The average lexicon size (dark yellow line) clearly shows how the agents locally introduce new words (leading to 13 different forms), before gradually converging on the optimal number of six words, one for each category (right y-axis). The blue line tracks the average number of forms per meaning in the population (right y-axis). In the phase in which many words are being invented, this number reaches its maximum, after which it gradually declines to a single form for each meaning. The green line shows the opposite, namely the average number of meanings per form (right y-axis). While the agents are still building up their ontologies, it can happen that two word forms get associated to the same meaning. As an effect of alignment, the meaning-per-form ratio gradually decreases to 1. 

\begin{figure}[h!]
\centering
\begin{subfigure}{\columnwidth}
        \includegraphics[width=.5\columnwidth]{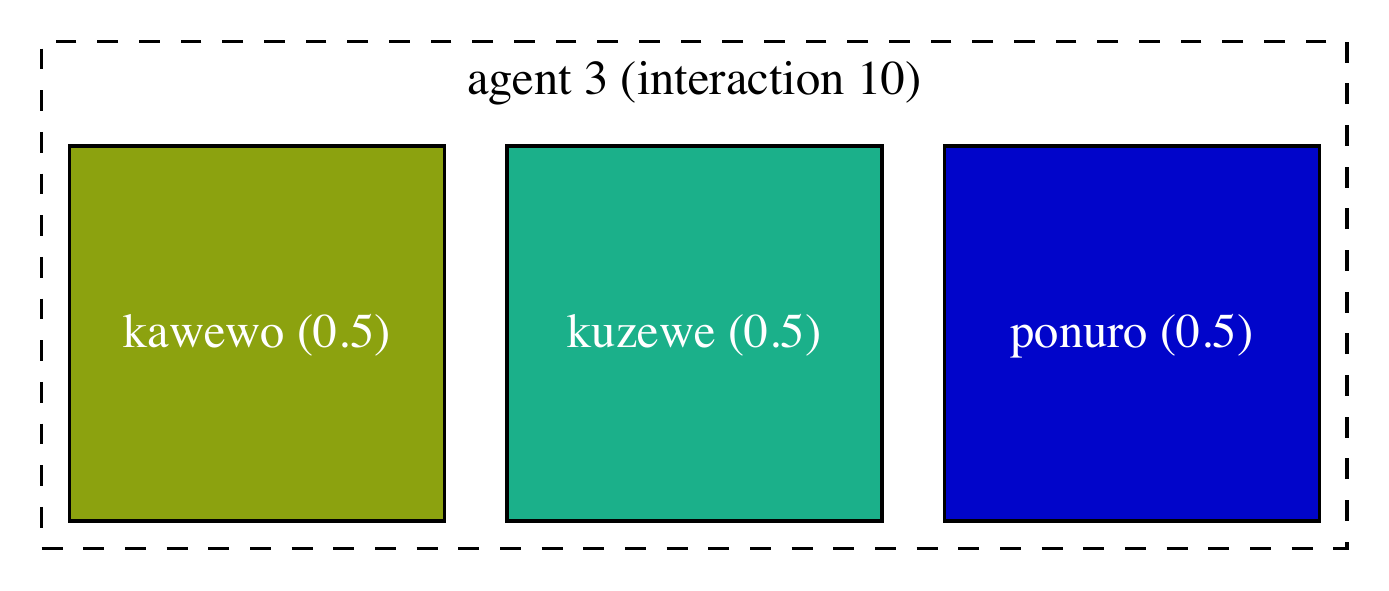}
    \end{subfigure}
    \begin{subfigure}{\columnwidth}
            \includegraphics[width=.65\columnwidth]{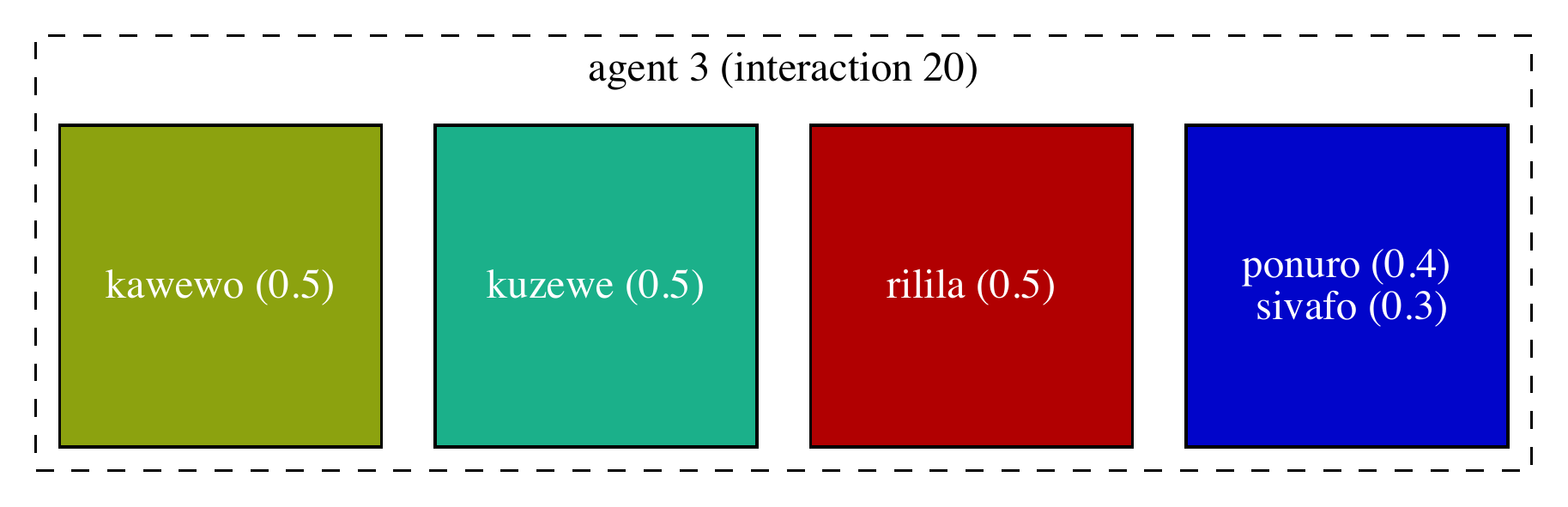}
    \end{subfigure}
   \begin{subfigure}{\columnwidth}
        \includegraphics[width=.8\columnwidth]{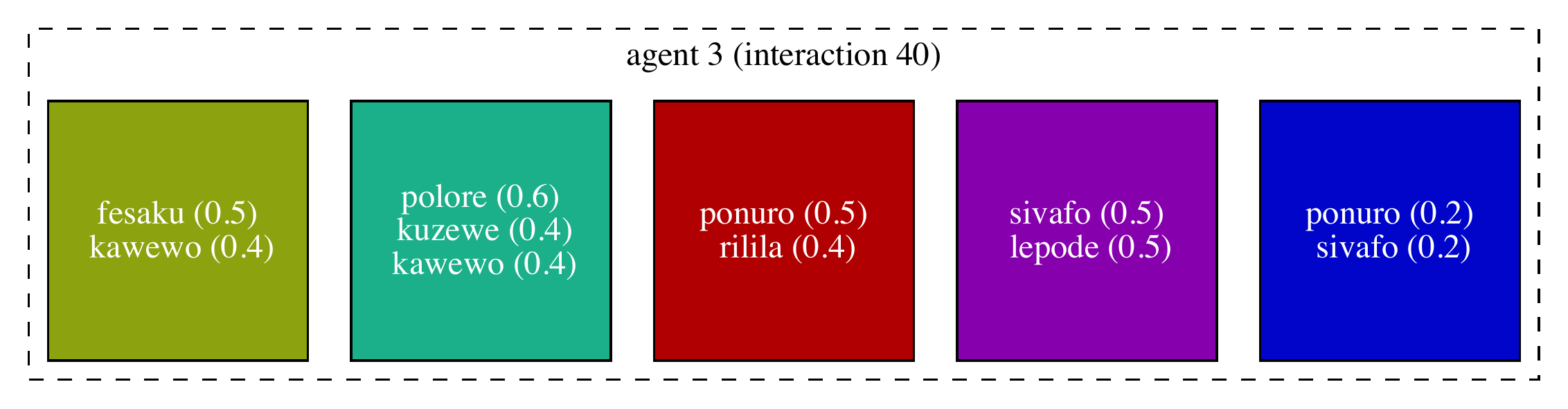}
    \end{subfigure}
    \begin{subfigure}{\columnwidth}
        \includegraphics[width=\columnwidth]{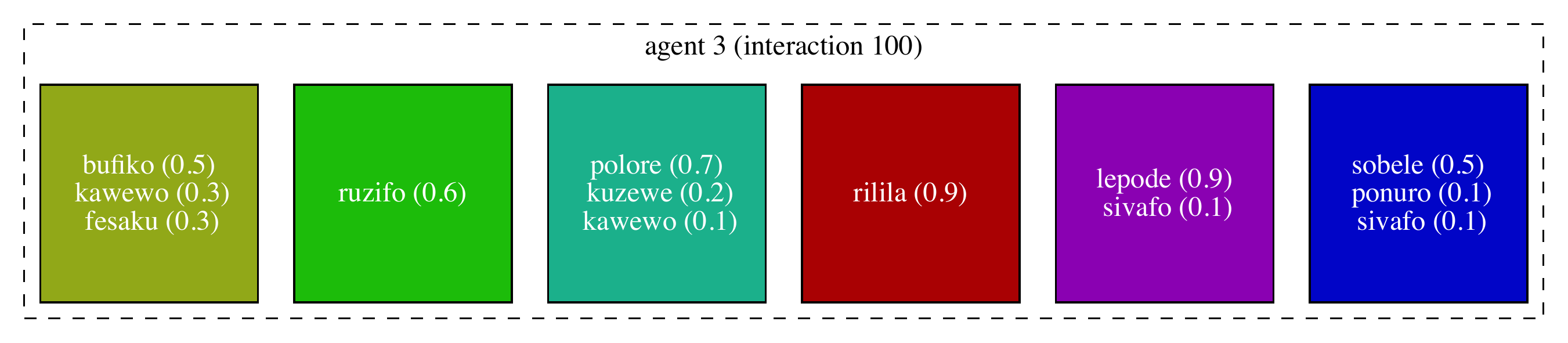}
    \end{subfigure}
    \begin{subfigure}{\columnwidth}
        \includegraphics[width=\columnwidth]{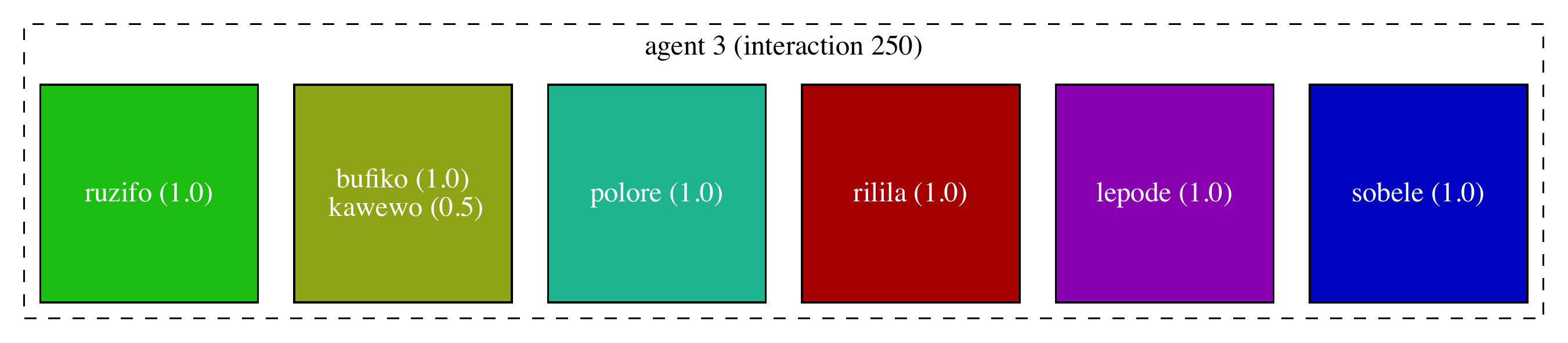}
    \end{subfigure}
\caption{A visualisation of agent 3's colour lexicon after 10, 20, 40, 100 and 250 games. Note that the word ``ponuro" was first exclusively used to refer to blueish objects (interaction 10) and was later also used to refer to reddish objects (interaction 40). In the end, the form did not survive, as the population converged on ``rilala" and ``sobele" for referring to reddish and blueish objects respectively (interaction 250).}
 \label{fig:colour-lexicons}
\end{figure}

A different kind of visualisation that was created using the monitoring system is presented in Figure \ref{fig:colour-lexicons}. Each row in the figure shows a snapshot of single agent's ontology of colour categories and their associated word forms at a specific point in time. In this case, the ontology and word forms of agent 3 are shown after 10, 20, 40, 100 and 250 interactions. We can see how the agent gradually distinguishes more colour categories, until he reaches the optimal number of six. We can also observe that after 100 interactions, the agent has learned multiple words for certain colour categories, but that most have already disappeared after 250 interactions. The rise and fall of the word ``ponuro'' is particularly interesting. It was first exclusively used to refer to blueish objects (interaction 10), was then also associated to the colour category used to refer to reddish objects (interaction 40), but dies out as the population has converged on the words ``ribala'' and ``sobele'' to refer to reddish and blueish objects respectively (interaction 250).

Babel's monitoring system can in real-time track, aggregate and visualise series of experiments that are run in parallel, and can easily be extended to record other experimental parameters or  measures.

\section{THE ROBOT INTERFACE: TECHNICAL SPECIFICATION}
\label{sec:robot-interface}

The robot interface is a newly developed part of the Babel software system, facilitating the implementation of processes that take place at the sensori-motor level of the semiotic cycle. It allows Babel users to seamlessly integrate the use of physical robot bodies in their language game experiments, by providing a hardware-independent interface to the functionality that is most frequently used in language games. This section first gives an overview of the general architecture of the robot interface, and then describes how it can be concretely used in combination with the Nao hardware that was employed in the experiment reported in the previous section.

\subsection{General architecture}

The robot interface standardises a number of core capabilities that can be exerted by a wide range of robotic platforms, abstracting away from their low-level implementation details. An overview of the most relevant capabilities, such as speaking, listening and pointing, is presented in Table \ref{tab:robot-interface}. 

\begin{table}
	\caption{Selected functions from the Babel robot interface API.}
	\begin{tabular}{lll}
		\toprule
		\textbf{Function} & \textbf{Arguments} & \textbf{Return value}\\
		\midrule
		\textsc{make-robot} & ip-address, port, type & robot-connection \\
		\midrule
		\textsc{observe-world} & robot-connection & world-model\\
		\midrule
		 \textsc{speak} & robot-connection, utterance & boolean \\
		 \textsc{hear} & robot-connection & perceived-utterance \\ 
		 \midrule 	
		  \textsc{point} & robot-connection, arm & boolean \\
		  \textsc{nod} & robot-connection & boolean \\
		  \textsc{shake-head} & robot-connection & boolean \\
		  \textsc{look-direction} & robot-connection, dir, angle & boolean \\
		\bottomrule
	\end{tabular}
	\label{tab:robot-interface}
\end{table}

In order to be able to use the robot interface, a robot-connection object of a specific type needs to be created first. This is done using the function \textsc{make-robot}, which takes an IP address, a port number and a type of robot (e.g. `nao') as input and returns a robot-connection object specialised towards this type of robot. Each of the available capabilities is then implemented as a Common Lisp generic function, with methods specialising on the subtype of the robot-connection object. This means for instance that when a certain capability is called with a robot-connection of type `nao' as its first argument, the call will automatically be dispatched to the method that implements this capability specifically for the Nao robot. A didactic example of how such a capability can be implemented is shown in Listing \ref{speak}. The example shows how the general \textsc{speak} capability is implemented as a generic function, while a call to this function with as first argument a connection of type `nao' will  automatically be routed to the method just below. This general architecture ensures that the robot interface is easily extensible, both in terms of adding additional functionality and in terms of extending the existing functionality to different robotic platforms.

\begin{lstlisting}[caption=Didactic example of the implementation of the speaking capability on a Nao robot., label=speak, basicstyle=\scriptsize\tt]
(defgeneric speak (robot-connection utterance)
  (:documentation ``The robot says the utterance.''))

(defmethod speak ((nao nao)
                  (utterance string))
  ``Sending the utterance to the Nao's speech endpoint, returning a boolean that indicates success or failure.''
  (rest (assoc :success
            (nao-send-data nao
                      :endpoint ``/speech/say''
                      :data `((speech . ,utterance))))))
\end{lstlisting}	

When setting up a grounded language game experiment like the one reported on in this paper, it suffices to create one robot-connection object per robot body, at the beginning of the experiment. At the start of each communicative interaction, the \textsc{embody} step  (see Section \ref{sec:multi}) will then assure that the speaker and hearer agents sense and act through the right robot body during this game, by associating them to one of these robot-connection objects. This avoids opening and closing a connection for every interaction.

\subsection{Using the Nao robot}

The experiment described earlier in this paper used the robot interface to play grounded colour naming games using two humanoid robots of the Nao type\footnote{\url{https://www.softbankrobotics.com/emea/en/nao}}. Nao robots run a GNU/linux-based operating system, called \textit{NaoQi OS}, and can be controlled from an external computer using the \textit{NaoQi framework}, available either as a C++ or a Python library. 

\begin{figure}
\includegraphics[width=\columnwidth]{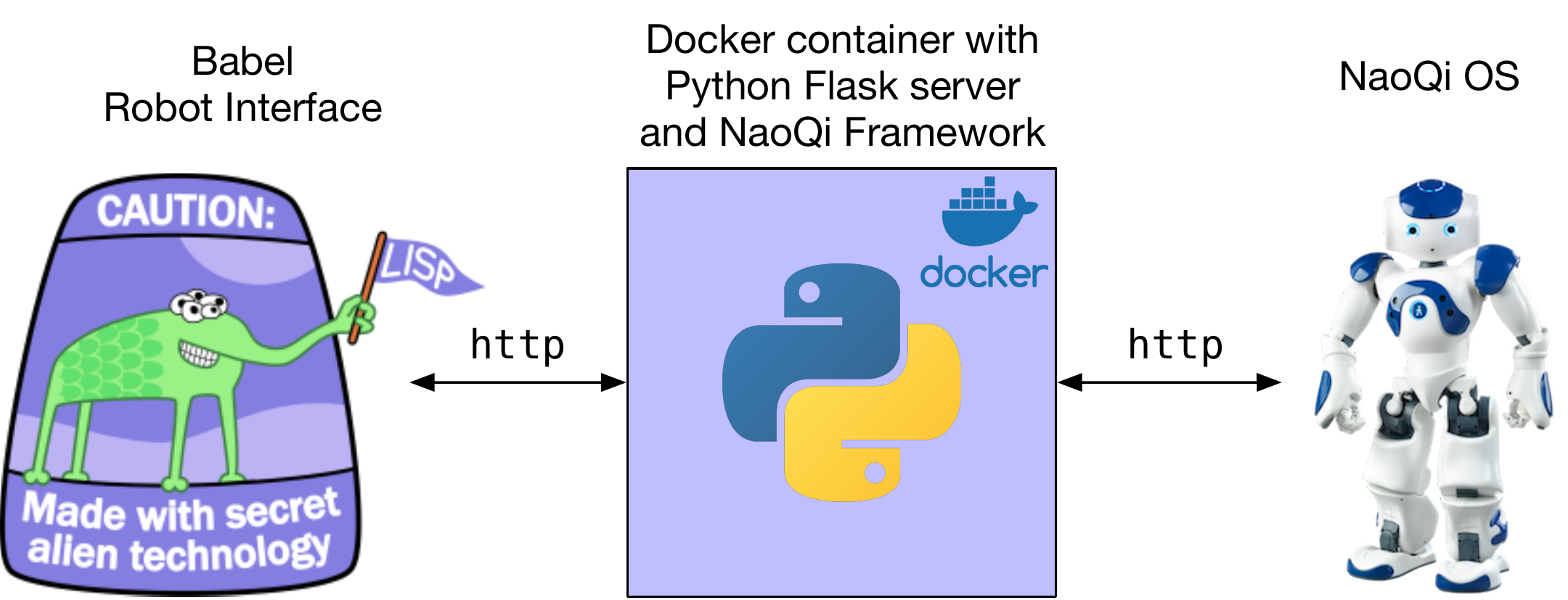}
\caption{When a function of the Babel robot interface API is called during an experiment, an HTTP POST request is send to a Python server running in a Docker container. The Python server then uses the NaoQi framework to communicate the request to the Nao robot.}
\label{fig:nao-interface}
\end{figure}

On the computer that runs the Babel software system, we set up a Docker container running a Python (Flask) server. This server exposes a RESTful API that continuously listens to HTTP requests, transforming them into concrete instructions that are passed to the right Nao robot using the Python version of the NaoQi framework. When  a function from the Babel robot-interface API is called during an experiment, an HTTP POST request containing the necessary information is sent to the Python server's endpoint that handles the capability associated to this function. Suppose for example that the function \textsc{speak} is called with as arguments a robot-connection object and an utterance. Babel's robot interface will then send an HTTP POST request containing a JSON object holding the IP address and port of the Nao associated to the robot-connection object, as well as the utterance to pronounce, to the \texttt{/speech/say} endpoint of the Python server running in the Docker container. The Python server will parse the request and call a function from the NaoQi framework that makes the robot say the utterance and will return a JSON object containing a key `success' with a boolean value. A visual depiction of this system architecture is shown in Figure~\ref{fig:nao-interface}. 

\section{CONCLUSION}
\label{sec:conclusion}

Grounded language game experiments form an excellent tool to study emergent communication and its underlying mechanisms. Setting up such experiments, in which a rich system for communicating about the real word emerges, requires implementing each process involved in the semiotic cycle, encompassing the sensori-motor, conceptual and language levels. This paper has introduced a high-level interface that allows making use of physical robots for operationalising the grounding processes on the sensori-motor level. This interface has been fully integrated into the Babel software system, which, as a result, now includes software modules that facilitate the implementation of all processes involved in the semiotic cycle. This paper has also presented a practical guide to using the Babel toolkit for setting up full-cycle experiments, taking the grounded colour naming game as didactic example.

\section*{ACKNOWLEDGEMENTS}
We would like to thank Remi van Trijp, Katie Mudd and Yannick Jadoul for their valuable comments on earlier versions of this paper. We are also grateful to the three anonymous reviewers of the AISB Symposium on Grounded Language Learning for Artificial Agents for their encouraging feedback and appreciation. This work was supported by the Research Foundation Flanders (FWO) through grants 1SB6219N and G0D6915N (CHIST-ERA ATLANTIS) and by the European Union's Horizon 2020 research and innovation programme under grant agreement No 732942 (ODYCCEUS). 

\bibliography{references}

\begin{thebibliography}{10}

\bibitem{baronchelli2006sharp}
Andrea Baronchelli, Maddalena Felici, Vittorio Loreto, Emanuele Caglioti, and
  Luc Steels, `Sharp transition towards shared vocabularies in multi-agent
  systems', {\em Journal of Statistical Mechanics: Theory and Experiment}, {\bf
  2006}(06),  P06014, (2006).

\bibitem{beuls2013agent}
Katrien Beuls and Luc Steels, `Agent-based models of strategies for the
  emergence and evolution of grammatical agreement', {\em PloS one}, {\bf
  8}(3),  e58960, (2013).

\bibitem{bleys2016language}
Joris Bleys, {\em Language strategies for the domain of colour}, Language
  Science Press, 2016.

\bibitem{bleys2009grounded}
Joris Bleys, Martin Loetzsch, Michael Spranger, and Luc Steels, `The grounded
  colour naming game', in {\em Proceedings of the 18th IEEE International
  Symposium on Robot and Human Interactive Communication (Ro-man 2009)},
  (2009).

\bibitem{opencv_library}
G.~Bradski, `{The OpenCV Library}', {\em Dr. Dobb's Journal of Software Tools},
  (2000).

\bibitem{cornudella2016role}
Miquel Cornudella, Thierry Poibeau, and Remi {van Trijp}, `The role of
  intrinsic motivation in artificial language emergence: a case study on
  colour', in {\em 26th International Conference on Computational Linguistics
  (COLING 2016)}, pp. 1646--1656, (2016).

\bibitem{de2006reach}
Bart De~Vylder and Karl Tuyls, `How to reach linguistic consensus: A proof of
  convergence for the naming game', {\em Journal of theoretical biology}, {\bf
  242}(4),  818--831, (2006).

\bibitem{garcia2016insight}
Emilia Garcia~Casademont and Luc Steels, `Insight grammar learning', {\em
  Journal of Cognitive Science}, {\bf 17}(1),  27--62, (2016).

\bibitem{lestrade_2016}
Sander Lestrade, `The emergence of argument marking', in {\em The Evolution of
  Language: Proceedings of the 11th International Conference (EVOLANGX11)},
  eds., S.G. Roberts, C.~Cuskley, L.~McCrohon, L.~Barcel\'{o}-Coblijn,
  O.~Feh\'{e}r, and T.~Verhoef. Online at
  \url{http://evolang.org/neworleans/papers/36.html}, (2016).

\bibitem{mole}
Sander Lestrade.
\newblock Mole: Modeling language evolution.
\newblock \url{https://CRAN.R- project.org/package=MoLE}, 2017.

\bibitem{loetzsch10why}
Martin Loetzsch and Michael Spranger, `Why robots?', in {\em Proceedings of the
  8th International Conference on the Evolution of Language (EVOLANG 8)}, eds.,
  Andrew~D.M. Smith, Marieke Shouwstra, Bart de~Boer, and Kenny Smith, pp.
  222--229. World Scientific, (2010).

\bibitem{loetzsch2008babel2}
Martin Loetzsch, Pieter Wellens, Joachim De~Beule, Joris Bleys, and Remi {van
  Trijp}, `The babel2 manual', {\em Technical Report AI-Memo 01-08}, (2008).

\bibitem{McIntyre:1998}
Angus McIntyre, `Babel: A testbed for research in origins of language', in {\em
  Proceedings of the 17th International Conference on Computational Linguistics
  - Volume 2}, COLING '98, pp. 830--834, Stroudsburg, PA, USA, (1998).
  Association for Computational Linguistics.

\bibitem{mordatch2017emergence}
Igor Mordatch and Pieter Abbeel.
\newblock Emergence of grounded compositional language in multi-agent
  populations, 2017.

\bibitem{oudeyer2006discovering}
Pierre-Yves Oudeyer and Fr\'ed\'eric Kaplan, `Discovering communication', {\em
  Connection Science}, {\bf 18}(2),  189--206, (2006).

\bibitem{pauw2012emergence}
Simon Pauw and Joseph Hilferty, `The emergence of quantifiers', {\em
  Experiments in cultural language evolution}, {\bf 3},  277, (2012).

\bibitem{puglisi2008cultural}
Andrea Puglisi, Andrea Baronchelli, and Vittorio Loreto, `Cultural route to the
  emergence of linguistic categories', {\em Proceedings of the National Academy
  of Sciences}, {\bf 105}(23),  7936--7940, (2008).

\bibitem{schueller2018}
William Schueller, {\em Active Control of Complexity Growth in Language Games},
  Ph.D.\ dissertation, University of Bordeaux, 2018.

\bibitem{spranger12perceptual}
Michael Spranger, Martin Loetzsch, and Luc Steels, `A perceptual system for
  language game experiments', in {\em Language Grounding in Robots}, eds., Luc
  Steels and Manfred Hild,  89--110, Springer, (2012).

\bibitem{spranger2012irl}
Michael Spranger, Simon Pauw, Martin Loetzsch, and Luc Steels, `{Open-ended
  Procedural Semantics}', in {\em {Language Grounding in Robots}}, eds.,
  L.~Steels and M.~Hild,  153--172, Springer, (2012).

\bibitem{spranger2012grammar}
Michael Spranger and Luc Steels, `{Emergent Functional Grammar for Space}', in
  {\em {E}xperiments in {C}ultural {L}anguage {E}volution}, ed., L.~Steels,
  number~3 in Advances in Interaction Studies,  207---232, John Benjamins,
  (2012).

\bibitem{steels1995self}
Luc Steels, `A self-organizing spatial vocabulary', {\em Artificial life}, {\bf
  2}(3),  319--332, (1995).

\bibitem{steels00e}
Luc Steels, `The emergence of grammar in communicating autonomous robotic
  agents', in {\em ECAI 2000: Proceedings of the 14th European Conference on
  Artificial Life}, ed., W.~Horn, pp. 764--769, Amsterdam, (August 2000). IOS
  Publishing.

\bibitem{steels2011design}
{\em Design Patterns in {Fluid Construction Grammar}}, ed., Luc Steels, John
  Benjamins, Amsterdam, 2011.

\bibitem{steels2012grounding}
Luc Steels, `Grounding language through evolutionary language games', in {\em
  Language Grounding in Robots},  1--22, Springer, (2012).

\bibitem{steels2017basics}
Luc Steels, `Basics of {Fluid Construction Grammar}', {\em Constructions and
  frames}, {\bf 9}(2),  178--225, (2017).

\bibitem{steels2005coordinating}
Luc Steels and Tony Belpaeme, `Coordinating perceptually grounded categories
  through language: A case study for colour', {\em Behavioral and brain
  sciences}, {\bf 28}(4),  469--488, (2005).

\bibitem{language-grounding-in-robots}
Luc Steels and Manfred Hild, {\em Language Grounding in Robots}, Springer,
  Berlin, 2012.

\bibitem{steels2000aibo}
Luc Steels and Fr\'ed\'eric Kaplan, `Aibo?s first words: The social learning of
  language and meaning', {\em Evolution of communication}, {\bf 4}(1),  3--32,
  (2000).

\bibitem{steels2010babel}
Luc Steels and Martin Loetzsch, `Babel: A tool for running experiments on the
  evolution of language', in {\em Evolution of communication and language in
  embodied agents},  307--313, Springer, (2010).

\bibitem{steels2012grounded}
Luc Steels and Martin Loetzsch, `The grounded naming game', {\em Experiments in
  cultural language evolution}, {\bf 3},  41--59, (2012).

\bibitem{steels2016boy}
Luc Steels, Martin Loetzsch, and Michael Spranger, `A boy named sue', {\em
  Belgian Journal of Linguistics}, {\bf 30}(1),  147--169, (2016).

\bibitem{steels1998spatially}
Luc Steels and Angus McIntyre, `Spatially distributed naming games', {\em
  Advances in complex systems}, {\bf 1}(04),  301--323, (1998).

\bibitem{van1999multiple}
Joris Van~Looveren, `Multiple word naming games', in {\em Proceedings of the
  11th Belgium-Netherlands Conference on Artificial Intelligence. Universiteit
  Maastricht, Maastricht, the Netherlands}, (1999).

\bibitem{van2008emergence}
Remi {van Trijp}, `The emergence of semantic roles in {Fluid Construction
  Grammar}', in {\em The Evolution Of Language},  346--353, World Scientific,
  (2008).

\bibitem{van2013linguistic}
Remi {van Trijp}, `Linguistic assessment criteria for explaining language
  change: A case study on syncretism in german definite articles', {\em
  Language Dynamics and Change}, {\bf 3}(1),  105--132, (2013).

\bibitem{vantrjip2016}
Remi {van}~Trijp, {\em The evolution of case grammar}, Language Science Press,
  2016.

\bibitem{netlogo}
Uri Wilensky.
\newblock Netlogo.
\newblock \url{http://ccl.northwestern.edu/netlogo/}, 1999.

\end{thebibliography}

\end{document}